\documentclass[11pt]{article}
\usepackage{acl}
\usepackage{times}
\usepackage{latexsym}
\usepackage{microtype}
\usepackage[utf8]{inputenc}
\usepackage{graphicx}
\usepackage{caption}
\usepackage{subcaption}
\usepackage{url}
\usepackage{booktabs}
\usepackage{multirow}
\usepackage{lscape}
\usepackage{comment}
\usepackage{colortbl}
\usepackage{pbox}
\usepackage{paralist}
\usepackage{amssymb}
\usepackage{amsmath} 
\usepackage{authblk}
\usepackage[noend]{algpseudocode}
\usepackage{textcomp}
\usepackage{pgfplots}
\pgfplotsset{width=8cm,compat=1.9}
\definecolor{light}{rgb}{0.5, 0.5, 0.5}

\usepackage[normalem]{ulem}
\useunder{\uline}{\ul}{}

\usepackage{floatrow}
\DeclareFloatFont{tiny}{\tiny}
\usepackage{tipa}

\usepackage[scaled=0.86]{helvet}

\usepackage{tikz}
\usepackage{collcell}
\usepackage{ragged2e}
\justifying

\newcounter{magicrownumbers}
\newcommand\rownumber{\stepcounter{magicrownumbers}\arabic{magicrownumbers}}

\newcommand*{\MinNumber}{0.2}%
\newcommand*{\MidNumber}{0.5}%
\newcommand*{\MaxNumber}{0.8}%
\newcommand{\ApplyGradient}[1]{%
        \ifdim #1 pt > \MidNumber pt
            \pgfmathsetmacro{\PercentColor}{max(min(100.0*(#1 - \MidNumber)/(\MaxNumber-\MidNumber),100.0),0.00)} %
            \hspace{-0.33em}\colorbox{green!\PercentColor!white}{#1}
        \else
            \pgfmathsetmacro{\PercentColor}{max(min(100.0*(\MidNumber - #1)/(\MidNumber-\MinNumber),100.0),0.00)} %
            \hspace{-0.33em}\colorbox{red!\PercentColor!white}{#1}
        \fi
}

\newcommand{\DEVELOPMENT}{0} 
\usepackage{ifthen}
\ifthenelse{\DEVELOPMENT = 1}{
	\newcommand{\tz}[1]{\textcolor{red}{\textbf{TZ:} #1}}		
	\newcommand{\pa}[1]{\textcolor{blue}{\textbf{PA:} #1}}
 	\newcommand{\oa}[1]{\textcolor{teal}{\textbf{OA:} #1}}
}{
\newcommand{\tz}[1]{}		
\newcommand{\pa}[1]{}
\newcommand{\oa}[1]{}
}

\title{Text or Image? What is More Important in Cross-Domain\\ Generalization Capabilities of Hate Meme Detection Models?}
\author[1]{Piush Aggarwal}
\author[2]{Jawar Mehrabanian}
\author[3]{Weigang Huang}
\author[4]{{\"O}zge Alacam}
\author[1]{Torsten Zesch} 
\affil[1]{CATALPA, FernUniversit{\"a}t in Hagen \textit {\{firstname.lastname@fernuni-hagen.de\}}}
\affil[2]{FernUniversit{\"a}t in Hagen \textit {\{jawar.mehrabanian@studium.fernuni-hagen.de\}}}
\affil[3]{Universit{\"a}t Duisburg-Essen \textit {\{weigang.huang@stud.uni-due.de\}}}
\affil[4]{LMU Munich and Universit{\"a}t Bielefeld \textit {\{oezge.alacam@uni-bielefeld.de\}}}

\date{}

\begin{document}

\maketitle

\begin{abstract}

This paper delves into the formidable challenge of cross-domain generalization in multimodal hate meme detection, presenting compelling findings. We provide enough pieces of evidence\footnote{Our code and dataset are released at \url{https://github.com/aggarwalpiush/HateDetection-TextVsVL}} supporting the hypothesis that only the textual component of hateful memes enables the existing multimodal classifier to generalize across different domains, while the image component proves highly sensitive to a specific training dataset. The evidence includes demonstrations showing that hate-text classifiers perform similarly to hate-meme classifiers in a zero-shot setting. Simultaneously, the introduction of captions generated from images of memes to the hate-meme classifier worsens performance by an average F1 of 0.02. Through blackbox explanations, we identify a substantial contribution of the text modality (average of 83\%), which diminishes with the introduction of meme's image captions (52\%). Additionally, our evaluation on a newly created confounder dataset reveals higher performance on text confounders as compared to image confounders with an average $\Delta$F1 of 0.18.
\end{abstract}

\section{Introduction}

Recently many hate-meme detection multimodal (MM) systems have been proposed, see \cite{https://doi.org/10.48550/arxiv.2205.04274} for a survey and \citep{kougia-pavlopoulos-2021-multimodal,aggarwal-etal-2021-vl, gold2021germemehate,https://doi.org/10.48550/arxiv.2012.08290,https://doi.org/10.48550/arxiv.2012.07788,https://doi.org/10.48550/arxiv.1908.03557,chen2020uniter} for individual contributions, but it is an ongoing concern that they do not generalize well in a cross-domain setting.
Possible causes are (i) the implicit knowledge captured by multi-modal hate messages (memes) \citep{Ma_2022_CVPR, https://doi.org/10.48550/arxiv.1910.03814,10.5555/3061053.3061172,10.1007/978-3-319-27433-1_4}, (ii) additional annotation noise in multi-modal settings \cite{oriol2019multimodal}, and (iii) more complex network architectures.

In this study, we explore the generalization capabilities of MM models for detecting hate memes. While previous studies  \citep{wang-etal-2020-cross-media, ma-etal-2021-effectiveness} support the significant role of image modality in other multimodal-based downstream tasks, however, in the case of meme classification, the meaning can only be correctly inferred from also looking at the image, so we find the analysis to be of special importance and worth replicating. Consequently, we initiate the evaluation of these models in settings outside their domain. We observe a significant decline in performance, with an average macro F1 score of $0.28$. 

\begin{figure}[t]
  \centering
  \includegraphics[scale=0.3]{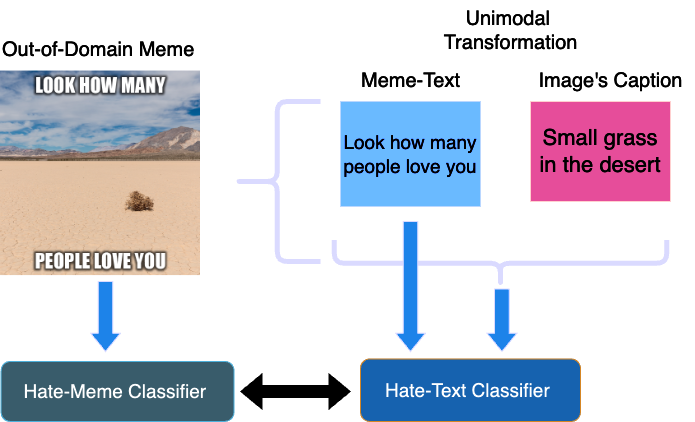}
  \caption{Illustration of our experimental arrangement for assessing the hate meme model's performance compared to unimodal text-based hate classifiers. The evaluation involves a test meme from a domain not included in the model's training data.}
  \label{fig:experimental-setup}
\end{figure}

\begin{table*}[t]
\centering
\small
\resizebox{\columnwidth}{!}{
	\begin{tabular}{llrrrr}
		\toprule
		\bf Name & \bf Reference & \bf \# Train/Dev/Test & \bf tokens & \bf \% hate & \bf Domain \\
		\midrule
		 \textsc{HarMeme} & \cite{pramanick-etal-2021-detecting} & 3.5k & 160k &  26.21 & Covid-19/US Election \\
		 \textsc{Mami} &  \cite{fersini2022semeval} &  10k &  590k &   50  & Misogynistic \\
		\textsc{FB} & \cite{https://doi.org/10.48550/arxiv.2005.04790} & 10k & 370k &   37.56 & \emph{mixed}  \\
		\bottomrule
	\end{tabular}
	}
	\caption{Properties of hate-meme datasets used in our study.} 
	\label{tab:datasets}
\end{table*}

\oa{I had problems in understanding the following sentence.}\pa{I try to make it more clear} We aim to tackle this issue by utilizing a text-only (unimodal) hate classifier, specifically crafted for the detection of hateful memes. Previous research \citep{nozza-2021-exposing, app10238614, Talat2018} demonstrates relatively higher generalization capabilities in the context of unimodal text-only hate. Our approach involves applying an unimodal transformation to memes by concatenating the text within the meme with a caption generated from the meme's image. Subsequently, we train a text-based classifier using a combination of nine diverse hate speech datasets and assess its performance on a transformed meme test set. We observe performance levels from our unimodal classifier that are comparable to those of MM models. In certain instances, our unimodal classifier even exhibits an improvement in performance, with an average F1 increase of $0.05$ compared to late-fusion-based MM models. The results make us infer that the text modality demonstrates superior generalizability compared to the image modality in detecting hateful memes. Figure~\ref{fig:experimental-setup} gives an overview of our experimental setup.

Additionally, we find that MM models behave differently than textual-based models. We retrain the MM models on hateful meme datasets which also include captions generated from the images available in the memes. Surprisingly, in comparison with existing models, in general, we find small performance drops (average $\Delta$F1 of $0.02$) in both in-domain as well as out-of-domain settings regardless of the presence of captions in the test sets. 

We explain the behavior of MM models by computing the contribution of text and image modality individually toward the prediction. We apply Shapley values \cite{parcalabescu-frank-2023-mm} to the features used in the models and average the final score for each modality (Section~\ref{shapcompute}). Our results indicate a substantial contribution ($83\%$) of textual modality by the models evaluated on all the datasets we have used in our study. Nevertheless, incorporating the image caption of the meme into the input data during the MM model training results in a decreased textual contribution of $52\%$. We believe that images in hateful memes are more like facilitators and provide context to the MM models.

To validate this, we compose a confounder dataset where we subset from the \textsc{HarMeme} and \textsc{FB} dataset \cite{pramanick-etal-2021-detecting, https://doi.org/10.48550/arxiv.2005.04790}, selecting 100 memes featuring celebrities or known figures such as \emph{Donald Trump, Nelson Mandela and Adolf Hitler}. 
We observe that MM models are sensitive to text confounders, while the prediction labels remain unchanged when the model is triggered with image confounders. (An average $\Delta$ F1 of $0.18$ is observed when the MM model is evaluated on Text and Image confounder sets).

Although, prior studies such as \citep{wang-etal-2020-cross-media, ma-etal-2021-effectiveness} have represented similar hypotheses. However, we find such studies for explicit types of downstream tasks.


In this paper, we present compelling evidence substantiating the hypothesis that the generalization of multimodal classifiers across diverse domains is primarily attributable to the textual component of hateful memes. Remarkably, our findings reveal a heightened sensitivity of the image part to the nuances of a specific training dataset. We believe we are the first to provide a thorough analysis supporting this idea, making our work unique in contributing to the field.

\section{Related Work}


\newcite{kirk-etal-2021-memes} demonstrate the high generalization behaviour of CLIP models \cite{Radford2019LanguageMA} when it is fine-tuned on the Hateful meme FB dataset \cite{https://doi.org/10.48550/arxiv.2005.04790} and tested on inhouse hate meme test set collected from pinterest\footnote{\url{https://www.pinterest.com/}}. However, their model is evaluated without using the meme's text which we believe provides significantly greater valid information for hateful meme detection. \newcite{cuo2022understanding} attempts to investigate the poor generalizability behavior of VL-models towards COVID-19-specific hate meme detection task. The application of the gradient-based explanation method demonstrates the significance of image modality is twice of textual one during predictions. Not specific to hate meme classification task, \newcite{Ma_2022_CVPR} evaluate the robustness of Visual-Linguistic transformers on missing modality datasets and found even poorer performance than uni-modal models and proposed a method that performs an optimal fusion of modalities which end up with better results. Error analysis of visio-linguistic models also indicates model bias \cite{10.1145/3485447.3512260}. While prior studies recommend investigating the contributions of each modality to model predictions to uncover the root cause of their limited generalization, these analyses tend to be overly specific, focusing solely on the in-house COVID-19 test set. Additionally, suggested methods like gradient-based explanation \cite{Selvaraju2019} are susceptible to deception through small input changes, as demonstrated in adversarial attacks \cite{parcalabescu-frank-2023-mm}. 

\section{Modality Contribution with Shapley Values}
\label{shapcompute}
Applying the method proposed by \citealp{parcalabescu-frank-2023-mm}, we attempt to investigate the modality contribution of existing hate meme detection models. There are multiple existing methods that can be used to estimate the importance of the model's features in the prediction process. Shapley values provide important ingredients for sample-based explanations that can be aggregated in a straightforward way into dataset-level explanations for machine
learning methods \cite{10.5555/3495724.3497168}. We calculate Shapley values for meme text tokens and image patches utilized in MM models during prediction. Each entity (token or patch) through its shapely value gauges its impact on the model prediction, such as the likelihood of image-sentence alignment. It can be positive (enhancing the model prediction), negative (diminishing it), or zero (no discernible effect).

\section{Datasets}

\subsection{Hateful Meme Datasets} 
In order to analyze the generalizability of available hate meme classifiers and modality contribution, we have used three benchmark datasets (see Table~\ref{tab:datasets}).

\paragraph{Kiela et al.} \cite{https://doi.org/10.48550/arxiv.2005.04790} (\textsc{Fb}) comprises 10,000 memes sourced from Getty images, semi-artificially annotated with benign confounders. It includes (i) \emph{multimodal hate} where both modalities possess benign confounders, (ii) \emph{unimodal hate} where at least one of the modalities is already hateful, (iii) \emph{benign image}, (iv)\emph{benign text} confounders and (v) \emph{random not-hateful} examples. The first four are labeled as \emph{hateful}, while the last is labeled as \emph{non-hateful}. The dataset is divided into 85\% training, 5\% development, and 10\% test sets, with balanced proportions for each meme variety in the development and test sets.

\paragraph{Pramanick et al.} \cite{pramanick-etal-2021-detecting} (\textsc{HarMeme}) consists of COVID-related memes from US social media, identified using keywords like \emph{Wuhan virus}, \emph{US election}, \emph{COVID vaccine}, \emph{work from home}, and \emph{Trump not wearing mask}. Unlike \cite{https://doi.org/10.48550/arxiv.2005.04790}, these memes are original, shared across social media, and their textual content is extracted using Google Vision API. The dataset is categorized into \emph{hateful} (including \emph{harmful} and \emph{partially harmful}) and \emph{non-hateful}, totaling 3,544 data points. The split for training, validation, and test sets is 85\%, 5\%, and 10\%, respectively.

\paragraph{Fersini et al.} \cite{fersini2022semeval} (\textsc{Mami}) focuses on SUBTASK-A, with memes labeled as \emph{misogynist} or \emph{non misogynist}. These are relabeled as \emph{hateful} and \emph{non-hateful} for consistency. The memes are collected from social media threads featuring women personalities such as Scarlett Johansson, Emilia Clarke, etc. as well as hashtags such as \texttt{\#girl}, \texttt{\#girlfriend}, \texttt{\#women}, \texttt{\#feminist}. Google Vision API is used for meme text extraction. With a balanced set of 10,000 instances, 10\% are used for both development and test sets, randomly stratified.

\subsection{Confounder Dataset}
\label{inhouseds}
\begin{figure}[t]
\centering
\resizebox{\columnwidth}{!}{
  \includegraphics[scale=0.2]{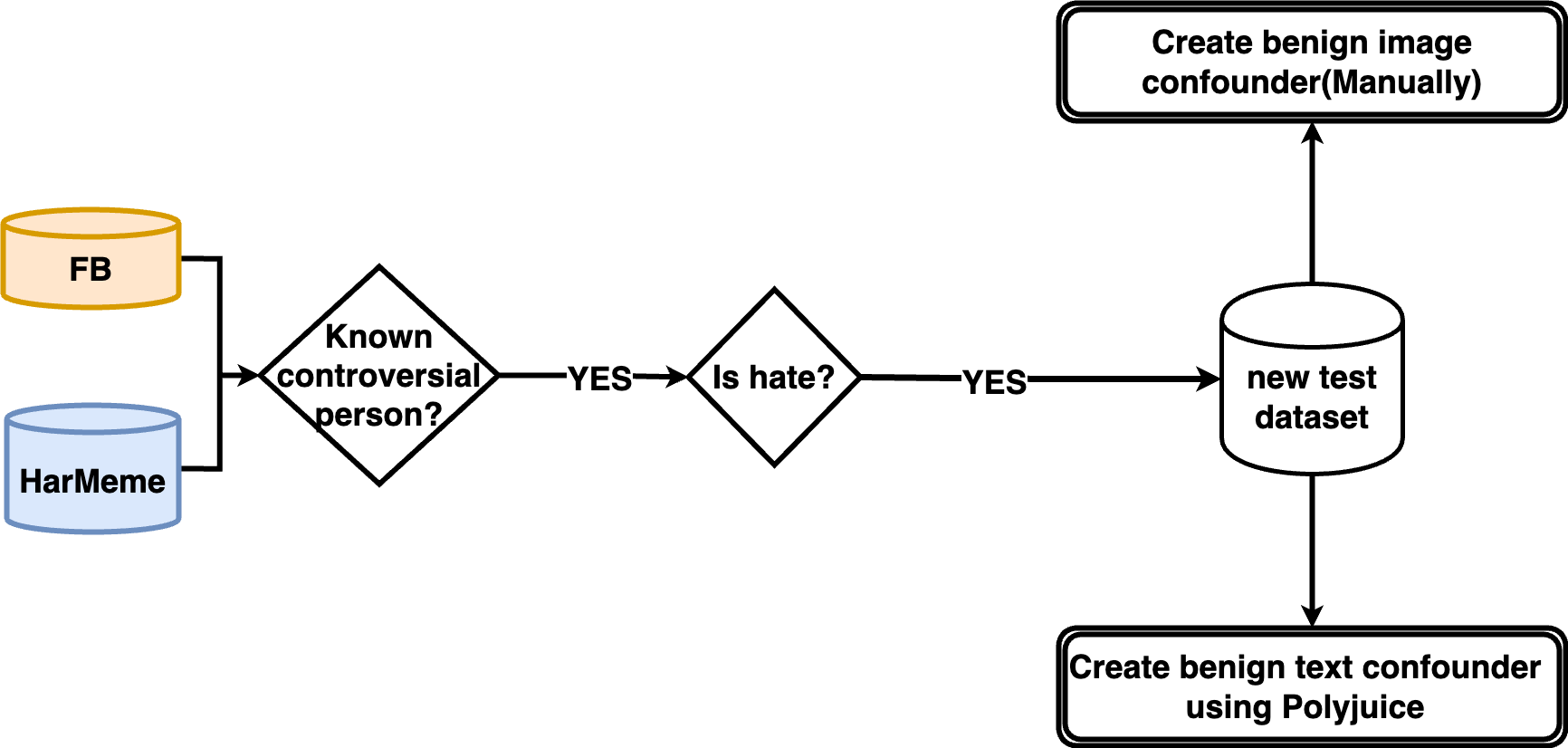}}
  \caption{A schematic showing the data collection process of our proposed dataset.}
  \label{fig:dataCollection}
\end{figure}

In order to validate the generalization capabilities of multimodal (MM) models for a specific modality, we create a tailored dataset for validation. We conducted an exhaustive search on the \textsc{FB} \cite{https://doi.org/10.48550/arxiv.2005.04790} dataset. A meticulous filtration process was implemented to exclude any instances featuring recognized celebrities or known figures such as \emph{Donald Trump, Nelson Mandela, and Adolf Hitler}. Subsequently, attention was directed towards memes labeled as \emph{hateful}. The selection was judiciously limited to a total of 100 figures, ensuring controlled and representative samples. The final stage of the methodology involved leveraging the identified set of hateful memes to construct a total of 100 benign images and text confounders. For image confounders, manual replacement of the celebrity figure with an analogous counterpart such as \emph{Anne Frank} with \emph{Adolf Hitler} (See Appendix~\ref{celebrityList} for complete list of the figures that were taken into account for the confounder dataset). Furthermore, to maintain simplicity and coherence for text confounders, the Polyjuice framework was incorporated \cite{wu-etal-2021-polyjuice}. It is a counterfactual generator, that is instrumental in facilitating control over the nature and positioning of perturbations in the textual content, enhancing the precision and consistency of the devised framework. Figure~\ref{fig:dataCollection} illustrates the data collection process for our proposed dataset.

\paragraph{Annotation process}
We recruited 12 annotators (university graduated volunteers and regular social media users) to read the introduction,  where the objective of the annotations along with the task is explained in detail (See Appendix~\ref{annotationProcess}). 
From each of the collected memes, we solicit various aspects for analysis. First and foremost, we inquire about the \emph{Image-text Relation}, seeking insights into the nuanced connection between the textual and visual components within a meme. Another crucial facet is the \emph{Modality towards Hate}, which serves as an evaluative measure for the modality of a meme that may convey hate or offensive content. For a more granular understanding, we introduce \emph{Decision Parts}, allowing annotators to pinpoint specific tokens or elements in the meme that contribute to its characterization as either hateful or non-hateful. To quantify the degree of hateful content, we employ a \emph{Hatefulness Score}, utilizing a scale that ranges from 0 to 5. A score of 0 denotes non-hateful content, while a score of 5 signifies highly hateful material. Additionally, annotators are prompted to provide a \emph{Confidence Score} reflecting their certainty regarding the accuracy of their judgments. This score operates on a scale from 0 (indicating a lack of confidence) to 5 (reflecting a high level of confidence). To maintain the integrity of the annotation process, we afford annotators the option to discard a sample, ensuring that only pertinent and valid memes\footnote{We offer annotators the option to exclude a meme if they lack sufficient knowledge to comprehend its content. Ultimately, we include only those memes that none of the annotators choose to discard.} are included in the analysis. We ended up with very good inter-rater agreement among the annotation with Krippendorff alpha \cite{krippendorff2011computing} as $0.8$. Furthermore, it is noteworthy that the average \emph{Confidence scores} is 4.38 out of 5 which shows very high confidence among the annotators.

\section{Experimental Models}
\label{expmodel}
\paragraph{Unimodal Hate Recognition} We use an online hate speech detection system called Perspective API\footnote{\url{https://www.perspectiveapi.com/}} which consists of multilingual BERT-based models trained on millions of comments from a variety of sources, including comments from online forums such as Wikipedia and The New York Times. These models are further distilled into single-language Convolutional Neural Networks (CNNs) for different languages. We also fine-tune BERT  \cite{https://doi.org/10.48550/arxiv.1810.04805} and SVM-based hate detection models on nine hate speech datasets which will be discussed in Section~\ref{textOnly}.

\paragraph{Multimodal Hate Recognition} Most of the promising studies on hate speech detection employ multi-modal based visual-linguistic pre-trained models \citep{chen2020uniter, https://doi.org/10.48550/arxiv.1908.03557,https://doi.org/10.48550/arxiv.2004.06165,https://doi.org/10.48550/arxiv.1908.08530,https://doi.org/10.48550/arxiv.1908.07490} which are originally designed to tackle basic visual-linguistic problems such as visual-question answering (VQA). These models caries semantic understanding between text and visual objects which makes them highly efficient for many downstream tasks. To analyze the vulnerability of the hate meme detection models, we investigate two early fusion and one late fusion-based multimodal (MM) models. \emph{VisualBert} \cite{https://doi.org/10.48550/arxiv.1908.03557}, an early fusion visual-linguistic transformer-based model, pre-trained on image caption as well as VQA datasets. We also investigate \emph{Uniter} model \cite{chen2020uniter} stands for UNiversal Image-TExt Representation which is also an early fusion visual-linguistic transformer-based model with additional pre-training with Visual Genome, Conceptual Captions, and SBU Captions. As the third MM model, we train a late-fusion ensemble model where we employ distinct extraction pipelines for image and text features. For image feature extraction, we utilize Resnet \cite{he2016deep}, a highly deep residual learning framework designed for generating image features.  To derive the text representation, we employ the widely-used RoBERTa model \cite{https://doi.org/10.48550/arxiv.1907.11692}. Subsequently, we concatenate the features from both modalities and feed them through a 128-layer feed-forward network with ReLU activation and a dropout rate of 0.2 to produce predictions. The model is trained for 30 epochs using the Adam optimizer \cite{https://doi.org/10.48550/arxiv.1412.6980}, with a learning rate of $10^{-5}$ and weight decay set to 0.1. This classifier is referred to as \emph{Rob+Resnet} for the purpose of illustration.

\paragraph{Image Caption Generation} We use \emph{ClipCap} \cite{https://doi.org/10.48550/arxiv.2111.09734}, which is based on Contrastive Language Image Pretraining (CLIP) \cite{https://doi.org/10.48550/arxiv.2103.00020} model to encode the image and pre-trained language model GPT-2 \cite{Radford2019LanguageMA} to decode a caption. We also use \emph{BLIP} \cite{li2022blip} which is a multimodal mixture of encoder-decoders optimized on three objectives during the pre-training process which include image-text contrastive loss, image-text matching loss, and language modeling loss. Unlike other models, it also performs caption bootstrapping in order to deal with noisy input data. We use both of these models in their default settings\footnote{\url{https://github.com/fkodom/clip-text-decoder}}.

\section{Experiments \& Results}

We conduct multiple sets of experiments in this study. Initially, we assess the cross-domain performance of hate-meme classifiers to gauge their generalization capabilities. Subsequently, we compare the performance of text-only hate classifiers on the textual component of memes with that of the hate-meme classifiers. We also assess the impact of captions generated from the image component of memes on text-only hate classifiers and hate meme detection models. Additionally, we compare the modality contribution from the blackbox explanations of the models with and without the introduction of captions. Finally, we apply the models to a confounder dataset to evaluate their sensitivity to a particular modality confounder set. 

\subsection{Generalization of Hate-meme Classifiers}
\label{genhatememe}











\begin{table}[t]
\centering
\small
\begin{tabular}{rlccc}
\toprule
& & \multicolumn{3}{c}{Test} \\

& \multicolumn{1}{l}{Train} & \rotatebox{90}{\textsc{HarMeme}} & \rotatebox{90}{\textsc{Mami}} & \rotatebox{90}{\textsc{FB}} \\ 
\midrule
\multirow{3}{*}{VisualBert}
& \textsc{HarMeme}    &  \cellcolor[gray]{.8}.80  & .40 & .48  \\

& \textsc{Mami}    &  .39 & \cellcolor[gray]{.8}.85 & .51  \\

  & \textsc{FB}    & .44 &  .60 & \cellcolor[gray]{.8}.66  \\

\midrule

\multirow{3}{*}{Uniter}
 &\textsc{HarMeme}    &  \cellcolor[gray]{.8}.79 & .45 & .48  \\

 & \textsc{Mami}    & .47 & \cellcolor[gray]{.8}.85 & .53  \\

  & \textsc{FB}    & .57 & .54 & \cellcolor[gray]{.8}.64  \\

\midrule
\multirow{3}{*}{Rob+Resnet}
 &\textsc{HarMeme}    &  \cellcolor[gray]{.8}.79 & .40 & .47  \\

 & \textsc{Mami}    & .39 & \cellcolor[gray]{.8}.83 & .45  \\

  & \textsc{FB}    & .41 & .49 & \cellcolor[gray]{.8}.62  \\
\bottomrule
\end{tabular}
\caption{F1(Macro) score of Hate-meme classifiers in cross-domain settings. \colorbox{gray}{Grey} highlighted values represent in-domain baselines.}
\label{tab:generalization}
\end{table}

To test the generalization capabilities of hate-meme classifiers, we fine-tune three state-of-the-art pre-trained models (VisualBert, Uniter and Rob+Resnet) on one datasets (train split) and test on the test splits of all three resulting in 9 train-test scenarios per model as can be seen in Table~\ref{tab:generalization}. Overall, we find huge performance drops across all the datasets for cross-domain testing. Since domains of \textsc{Harmeme} and \textsc{Mami} are exclusive, we encounter symmetry among each other (F1 of .398 and .393 for VisualBert and .453 and .467 for Uniter). On the hand, for \textsc{FB}, as there is no specific domain, we find relatively less decrement (however it is still huge) in the F1 scores. The results clearly infer a lack of generalization capabilities among these models.

\subsection{Zero-shot Text-only Classifiers}
\label{textOnly}
\begin{table}[t]
\centering
\small
	\begin{tabular}{lrrr}
		\toprule
		 \bf Reference & \bf \# Posts & \bf tokens & \bf \% hate \\
		\midrule
	\cite{hateoffensive} & 25K & 245K &   6  \\
		 \cite{Mollas_2022} & 1K & 14K &  43  \\
	  \cite{kennedy2022introducing} &  28K &  411K &   15     \\
   \cite{https://doi.org/10.48550/arxiv.1809.04444} &  10K &  169K &   11  \\
   \cite{https://doi.org/10.48550/arxiv.2108.05927} &  7K &  174K &   36   \\
      \cite{Basile2019SemEval2019T5} &  13K &  254K &   4   \\
       \cite{samoshyn_2020} &  2K &  38K &   48  \\
       \cite{waseem-hovy:2016:N16-2} &  17K &  131K &   32  \\
       \cite{waseem:2016:NLPandCSS} &  4K &  31K &   16   \\
       \addlinespace[1 mm]
       \cline{1-4}
       \addlinespace[1 mm]
       Total & 107K & 1467K & 23 \\
		\bottomrule
	\end{tabular}
	\caption{Hatespeech datasets used to train the hate-text classifiers. For all datasets, the collection is based on hate slurs matching, therefore all of them consist \emph{mixed} domains.}
	\label{textualdataset}
\end{table}

\begin{table}[t]
\centering
\small
\begin{tabular}{llcccr}
\toprule
& &  \multicolumn{3}{c}{Testset} &\\

&\multicolumn{1}{l}{Model} & \rotatebox{90}{\textsc{HarMeme}} & \rotatebox{90}{\textsc{mami}} & \rotatebox{90}{\textsc{FB}} &\\ 
\midrule
\multirow{3}{*}{\pbox{20cm}{Image +\\Text}} 
& VisualBert & \ApplyGradient{.44} & \ApplyGradient{.60} & \ApplyGradient{.51} & \rownumber \\
& Uniter     & \ApplyGradient{.57} & \ApplyGradient{.54} & \ApplyGradient{.53} & \rownumber\\
& Rob+Resnet & \ApplyGradient{.47} & \ApplyGradient{.45} & \ApplyGradient{.49} & \rownumber\\

\midrule
\multirow{3}{*}{Text}
& Perspective API  & \ApplyGradient{.45} & \ApplyGradient{.52} & \ApplyGradient{.49} & \rownumber\\
& BERT             & \ApplyGradient{.48} & \ApplyGradient{.53} & \ApplyGradient{.52} & \rownumber\\
& SVM              &  \ApplyGradient{.45} & \ApplyGradient{.45} & \ApplyGradient{.41} & \rownumber\\
\midrule
\multirow{3}{*}{\pbox{20cm}{Text +  \\ caption \\(ClipCap)}}
& Perspective API  & \ApplyGradient{.50} & \ApplyGradient{.52} & \ApplyGradient{.53} & \rownumber\\
& BERT             & \ApplyGradient{.50} & \ApplyGradient{.54} & \ApplyGradient{.52} & \rownumber\\
& SVM              & \ApplyGradient{.47} & \ApplyGradient{.45} & \ApplyGradient{.43} & \rownumber\\
\midrule
\multirow{3}{*}{\pbox{20cm}{Text + \\ caption \\(BLIP)}}
& Perspective API  & \ApplyGradient{.50} & \ApplyGradient{.53} & \ApplyGradient{.53} & \rownumber\\
& BERT             & \ApplyGradient{.51} & \ApplyGradient{.53} &  \ApplyGradient{.53}& \rownumber\\
& SVM              & \ApplyGradient{.46} & \ApplyGradient{.44} & \ApplyGradient{.43} & \rownumber\\
\bottomrule

\end{tabular}
\caption{Hate-meme vs.\ Hate-text Classifiers F1 Performance on cross-domain data. For Hate-meme classifiers, we indicate the best F1 value among the two training sets. A color gradient ranging from \colorbox{red}{red} to \colorbox{green}{green} is employed to emphasize the transition from lower to higher F1 values, respectively.}
\label{rivals}
\end{table}

\begin{table*}[t]
\centering
\small
\begin{tabular}{rlcccccc}
\toprule
& & \multicolumn{3}{c}{Test (Meme text)} & \multicolumn{3}{c}{Test (with Caption)}\\

& \multicolumn{1}{l}{Train} & \rotatebox{90}{\textsc{HarMeme}} & \rotatebox{90}{\textsc{Mami}} & \rotatebox{90}{\textsc{FB}} & \rotatebox{90}{\textsc{HarMeme}} & \rotatebox{90}{\textsc{Mami}} & \rotatebox{90}{\textsc{FB}} \\ 
\midrule
\multirow{3}{*}{Meme text}
& \textsc{HarMeme}    &  \ApplyGradient{.79}  & \ApplyGradient{.40} & \ApplyGradient{.47} &  \ApplyGradient{.65}  & \ApplyGradient{.51} & \ApplyGradient{.55} \\

& \textsc{Mami}    &  \ApplyGradient{.39} & \ApplyGradient{.83} & \ApplyGradient{.45}  &  \ApplyGradient{.40}  & \ApplyGradient{.81} & \ApplyGradient{.47}\\

  & \textsc{FB}    & \ApplyGradient{.41} &  \ApplyGradient{.49} & \ApplyGradient{.62}  &  \ApplyGradient{.35}  & \ApplyGradient{.49} & \ApplyGradient{.56}\\

\midrule

\multirow{3}{*}{With Caption}
 &\textsc{HarMeme}    &  \ApplyGradient{.77} & \ApplyGradient{.41} & \ApplyGradient{.46}  &  \ApplyGradient{.65}  & \ApplyGradient{.52} & \ApplyGradient{.53}\\

 & \textsc{Mami}    & \ApplyGradient{.39}  & \ApplyGradient{.77} & \ApplyGradient{.42} &  \ApplyGradient{.39}  & \ApplyGradient{.78} & \ApplyGradient{.44} \\

  & \textsc{FB}    & \ApplyGradient{.41} & \ApplyGradient{.50} & \ApplyGradient{.49} &  \ApplyGradient{.34}  & \ApplyGradient{.49} & \ApplyGradient{.48} \\

\bottomrule
\end{tabular}
\caption{F1(Macro) score of Roberta+Resnet based Hate-meme classifiers when trained with image caption.}
\label{captioneffect}
\end{table*}

We now compare the multimodal hate-meme classifiers to unimodal text-only classifiers.
For that purpose, we train two text-only classifiers (SVM and BERT) on a large collection of hate speech datasets (see Table~\ref{textualdataset}). Overall, we use around $0.1$ Million posts having $1.4$ Million tokens out of which 23\% posts are hateful. In the case of SVM, for tokenization and feature extraction, we use ArkTokenizer and fastext embeddings respectively. In the case of BERT, we follow the uncased-large model\footnote{\url{https://huggingface.co/bert-large-uncased}} for fine-tuning.
We also use the hate speech classifier as provided by the Perspective API\footnote{\url{https://www.perspectiveapi.com/}} which outputs a toxicity score for a given text.
A toxicity score greater than 50\% is considered hate otherwise non-hate.

Table~\ref{rivals} compares the zero-shot domain transfer results of hate-meme and hate-text classifiers. We encounter a close resemblance between them in their performances. Among cases where the text-only classifier is applied only on meme text, BERT model performance is superior to the rest of the two with an average F1 score of .51 followed by Perspective API (F1 of .59) (depicted in Table's line 4 and 5). We observe a similar performance by multimodal hateful meme classifiers (average F1 of .52 for \emph{VisualBert} and .55 for \emph{Uniter} and .47 for \emph{Rob+Resnet}) (see line 1, 2 and 3). 

With the quite good performance of the text-only classifiers, it might be worthwhile trying to extract the semantics of the image as text. For this purpose, we append captions generated by caption models (see Section~\ref{expmodel}) along with meme text and input to the hate-text classifier that we have trained on multiple corpora (as described in Section~\ref{textOnly}). Table~\ref{rivals} illustrates the performance of \emph{ClipCap} and \emph{BLIP} models. Compared with hate-meme and hate-text classifiers, we find a slight improvement in BERT with an average F1 of .52 which is 1 point higher than BERT tested only on meme text (depicted in line 8). 
However, it is 3 points lower than the Uniter model. Notably, Perspective API exhibits an improvement in performance, with an average F1 increase of $0.05$ compared to \emph{Rob+Resnet} model (depicted in line 10). This outcome suggests that classifier generalization is predominantly influenced by the textual modality. This pattern further implies a potential bias towards textual elements in meme data, leading to limitations in the ability of the multimodal model to integrate image meaning for this particular task. Mann-Whitney U Test shows that the results are statistically significant with $p<0.05$.

\subsection{Impact of Captions} 
In this section, we illustrate the effect of incorporating the captions in the training.
During the training process for each of the hateful meme classifiers, we incorporate image captions generated using a \emph{BLIP} model into the \emph{Rob+Resnet}. We then assess the performance of the resulting model that includes captions in comparison to its original counterpart. Our evaluation is conducted both on (i) the original test set and (ii) plus with captions. In Table~\ref{captioneffect}, it is evident that when the models trained with including captions perform poorly in both in-domain and out-of-domain testing scenarios, regardless of the presence or absence of captions in the test sets. A plausible explanation for this phenomenon could be the neutralization of contextual nuances introduced by the supplementary captions in the meme's text.  \oa{which will be discussed in the Discussion}\oa{What about the MAMI set, it also needs to be clarified, looks interesting}However, we also see a performance increase (average $\Delta$ F1 of 0.09) in the case of the model trained on \textsc{Harmeme} dataset when tested on an out-of-domain test set with concatenated captions. One potential explanation for this behavior could be attributed to the high resolution of the original images in this dataset, marked by an average bit depth of 43.90, a notable contrast to other datasets, with bit depths of 9.54 for the \textsc{FB} and 4.30 for the \textsc{mami} dataset \cite{10.1145/3543507.3583356}.

\begin{figure}[!ht] \centering
\begin{tabular}[c]{@{}c@{}}
     \begin{subfigure}[c]{.30\linewidth}
         \centering
         \includegraphics[width=\linewidth]{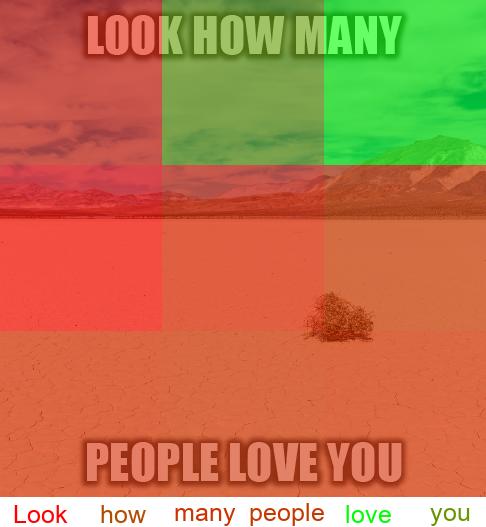}
         \caption{FB \\ TS=0.69 \\  F1=0.62}
         \label{fbo}
     \end{subfigure}
\hfill
    \begin{subfigure}[c]{.30\linewidth}
         \centering
         \includegraphics[width=\linewidth]{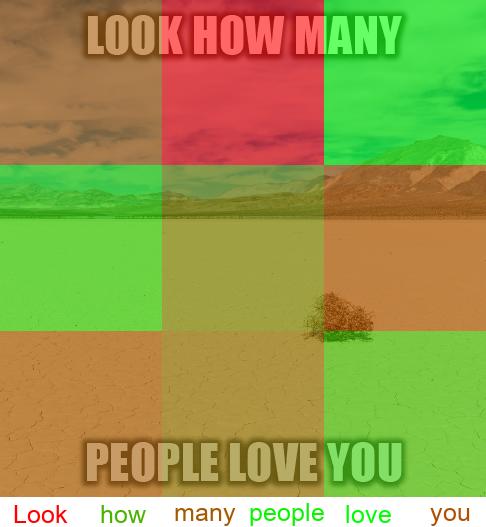}
          \caption{Harmeme \\ TS=0.85 \\  F1=0.79}
         \label{Harmemeo}
     \end{subfigure}
\hfill
    \begin{subfigure}[c]{.30\linewidth}
         \centering
         \includegraphics[width=\linewidth]{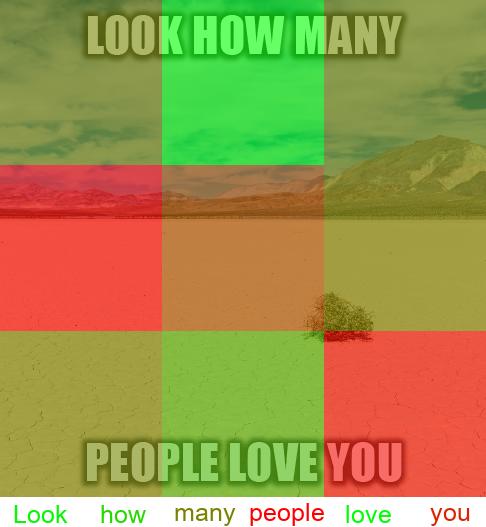}
          \caption{Mami\\ TS=0.95 \\  F1=0.83}
         \label{mamio}
     \end{subfigure}\\
\noalign{\bigskip}%
    \begin{subfigure}[c]{.30\linewidth}
         \centering
         \includegraphics[width=\linewidth,page=2]{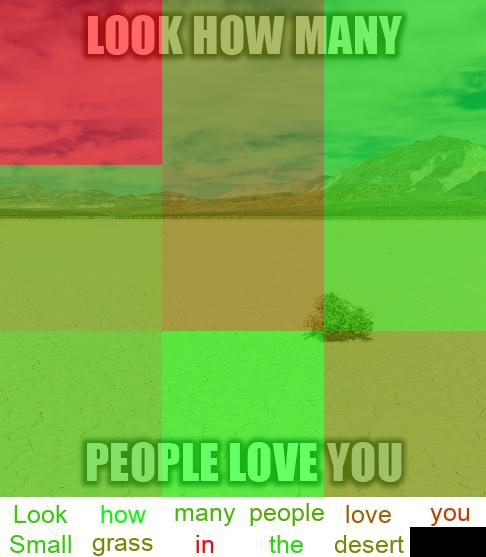}
          \caption{FB (+C)\\ TS=0.12 \\  F1=0.48}
         \label{fbc}
     \end{subfigure}

    \hfill
    \begin{subfigure}[c]{.30\linewidth}
         \centering
         \includegraphics[width=\linewidth]{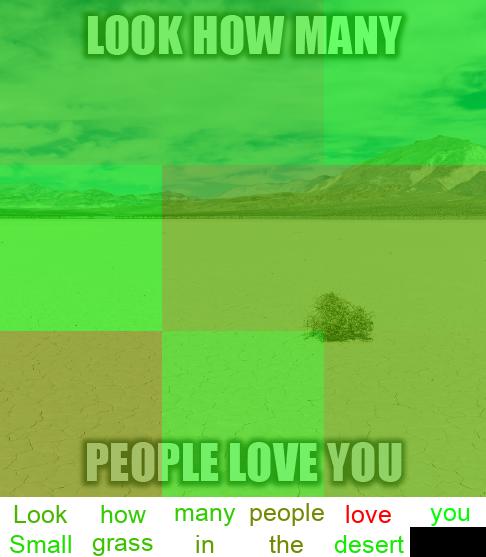}
          \caption{Harmeme (+C)\\ TS=0.67 \\  F1=0.65}
         \label{harmemeic}
     \end{subfigure}
     \hfill
    \begin{subfigure}[c]{.30\linewidth}
         \centering
         \includegraphics[width=\linewidth]{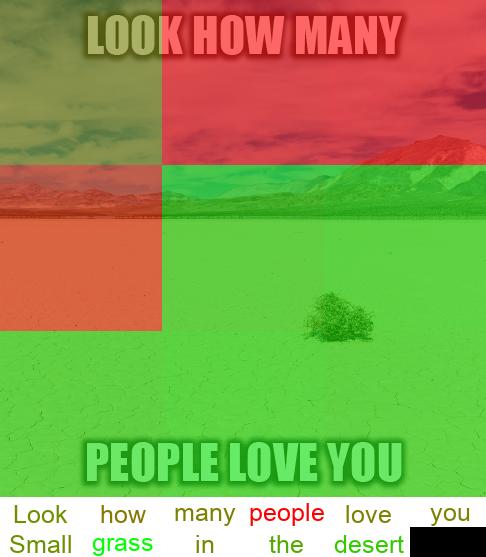}
          \caption{Mami (+C) \\ TS=0.76 \\  F1=0.78}
         \label{mamiic}
     \end{subfigure}\\
        
     \end{tabular}
        \caption{Example of Modality Contrbution of Rob+Resnet based hate meme detection model when trained on different hateful meme datasets. Here notation (+C) refers additional caption used in model training. \colorbox{red}{RED} and \colorbox{green}{GREEN} colour illustrate low and high contribution respectively.}
        \label{mmshap_testimg}
\end{figure}

\subsection{Impact of Modality}

\begin{table*}[t]
\centering
\small
\begin{tabular}{rlcccc}
\toprule
& & \multicolumn{2}{c}{Test (Meme text)} & \multicolumn{2}{c}{(with Cap+Celeb)}\\

& \multicolumn{1}{l}{Train} & I & T  &  $I^+$ & $T^+$\\ 
\midrule
\multirow{3}{*}{Meme text}
& \textsc{HarMeme}    &  \ApplyGradient{.42}  & \ApplyGradient{.45} & \ApplyGradient{.46}  &  \ApplyGradient{.44}   \\

& \textsc{Mami}    &  \ApplyGradient{.17} & \ApplyGradient{.39} & \ApplyGradient{.19} &  \ApplyGradient{.39}  \\

  & \textsc{FB}    & \ApplyGradient{.43} &  \ApplyGradient{.75} & \ApplyGradient{.41} &  \ApplyGradient{.72} \\

\midrule

\multirow{3}{*}{With Caption}
 &\textsc{HarMeme}    &  \ApplyGradient{.27} & \ApplyGradient{.39} & \ApplyGradient{.34}  &  \ApplyGradient{.43}  \\

 & \textsc{Mami}    & \ApplyGradient{.10}  & \ApplyGradient{.34} & \ApplyGradient{.17} &  \ApplyGradient{.42}  \\

  & \textsc{FB}    & \ApplyGradient{.33} & \ApplyGradient{.53} & \ApplyGradient{.35} &  \ApplyGradient{.54} \\

\bottomrule
\end{tabular}
\caption{F1(Macro) score of Roberta+Resnet based Hate-meme classifiers on Confounder datasets (T: Text Confounders, I: Image Confounders).}
\label{confounderResults}
\end{table*}

\paragraph{Shapley Values Computation} 
To calculate modality contribution, we determine Shapley values for feature maps, which are utilized by MM models for prediction. To achieve this, we generated patches of meme images such that each text token will be generally represented in a patch. From the existing set of image patches and text tokens (entities), we selected a subset and masked the remaining entities in the set. The determination of the number of subsets was influenced by the Monte Carlo approximation method. The Shapley value for each entity is computed by subtracting the model's output while it is present from that while it is absent. The resulting value was normalized considering the possible combinations of subsets. Ultimately, to compute the Shapley values for text contributions, the result outcomes of textual tokens are summed and normalized. The following algorithm delineates the process of generating modality scores.

\begin{flushleft}
\textbf{Input:} \texttt{Meme image $I$, Meme text $T$, model $f$, random number $P$, Shapley Value $\phi$, Text contribution score $TS$} \\
\text{image patches} $ {I_p = \lceil\sqrt{\text{len}(T)}\rceil} ^2$
\end{flushleft}
\begin{algorithmic}[1]
\ForAll{\texttt{$t$ $\epsilon$ ${t_1, . . . , (I_p + T)}$}} 

\ForAll{\texttt{$i$ $\epsilon$ ${1, . . . , 2*P+1}$}}
\State choose subset $S  \subseteq (I_p + T)$ \text{where} $len(S) = i$ \text{and} $t \notin
 S$ 
\State $\phi(t)$ = $\sum_{S,t}$ $\frac{f(S+t) - f(S)}{\gamma}$ \text{where} $\gamma$ \text{is normalizing factor}
\EndFor
\EndFor
\State $\phi(T) = \sum_{n=1}^{len(T)} \phi(t_n)$
\State $\phi(I) = \sum_{n=1}^{len(I_p)} \phi(t_n)$
\begin{flushleft}
\Return $TS = \frac{\phi(T)}{\phi(T) + \phi(I_p)}$
\end{flushleft}
\end{algorithmic}

\vspace{2mm}

\paragraph{Explanation of Hate Meme Detection Models} We calculate the contributions of modalities toward predictions of late fusion ensemble based \emph{Rob+Resnet} model using Shapley values. The classifiers are trained on each of the datasets with and without captions concatenated with the meme's text. To illustrate, Figure~\ref{mmshap_testimg} shows  Shapely values on a meme example for different models. The colours \colorbox{red}{RED} and \colorbox{green}{GREEN} indicate low and high contributions, respectively. In addition, the Text Contribution score (TS) as well as the F1 score evaluated on the in-domain evaluation set is provided in the caption of each of the subfigures. Image modality contribution (IS) can be computed using $IS = 1 - TS$.
We find that text contribution is quite higher ($TS >> IS$) for all the models (average TS of $.83$). When a caption is added to the text, the contribution score of the text modality decreases to $.52$. With this we infer that adding captions to memes strengthens the focus on the image modality. We observed that when we include the image caption along with the meme's text, the models establish a correlation between the caption and the meme's image. In such cases, the models tend to focus on the image’s information related to the image caption, a behavior not exhibited when the caption is absent. It infers that the meme text inherently carries a more potent message of hatefulness, which is mitigated by the inclusion of image captions. Nevertheless, it's important to highlight that the F1 score also decreases when captions are introduced to meme text.  This might also mean that to the existing models, images in hateful memes are more like facilitators and provide context to the models. As an example, in Figure~\ref{mmshap_testimg} we see that the dominancy of image patches is much higher for models trained along with captions. Similarly, we also see less dominancy of important hate context tokens such as \emph{LOVE} \cite{10.1145/3270101.3270103, aggarwal-zesch-2022-analyzing} in this case.

\oa{the insights obtained from the analysis of modality-specific contributions with the diminishing effect of including captions indicate that the current multimodal models are focused on finding the alignment between the image and text tokens at a concrete level. Adding captions generated by other multimodal models misdirects the attention to those low-level alignments while text-image alignment in hate-speech usually occurs at a more abstract (metaphorical) level. It is very clear that the meaning of the image could be extracted and incorporated at a higher level, where the current models and training regimes are falling short of addressing (my formulation is very poorly written, but my interpretation of the results is something like this, this could be mentioned in the Discussion or Conclusion session.}

\subsection{Classifiers on Confounder Dataset} In Section~\ref{inhouseds}, we elaborate on the composition of the confounder dataset. We divide it into two subsets. The first subset is termed the text confounder set (T), wherein meme instances are categorized based on images resembling those in hateful memes. Similarly, the second subset is designated as the image confounder set (I), where meme instances are categorized based on text resembling that found in hateful memes. In addition, we also concatenate the textual component of these sets with the image's caption and names of the celebrities available in the image and called them extended sets ($T^+$ and $I^+$). In this way, we have four evaluation sets to assess hate meme classifiers.

We evaluate \emph{Rob+Resnet} classifier which is already trained on the original hateful meme datasets and also in concatenation with captions. Table~\ref{confounderResults} illustrates the classifier's performance in terms of F1 (macro) scores. Overall, the performance on T is notably higher than that on the I across all variants of models. However, this difference is quite small in the case of \textsc{Harmeme} dataset (the average $\Delta$ F1 is 0.26, 0.23, and 0.08 for \textsc{FB}, \textsc{mami} and \textsc{harmeme} respectively). A similar trend is observed in the case of extended sets. Overall there is $\Delta$ F1 of 0.18 is observed which illustrates that the classifier is highly sensitive to memes undergoing changes in text while maintaining the same image, a sensitivity not observed in the other modality. Similar to the observations in Table~\ref{captioneffect}, the addition of captions to the meme's text significantly reduces performance for both the image and text confounder sets. This further adds evidence of the importance of the textual component of memes for hate detection models.

\oa{I think we have two nice opportunities to provide Error Analysis sections, one after section 6.1 or 6.2  (because I would like to learn more about why MAMI is so different than others not effecting at all from caption inclusion, or what makes HARMEME easier than FB, by looking at the wrongly classified samples, and their captions etc. And the other one after 6.5 providing some samples (in the appendix) where the image is contributive, where the text is contributive or when they have similar weights . These two deeper look would strengthen the paper a lot, provide valuable insights for some discussion! It just came to my mind yet (a little late), but training on the combined harmeme+fb+mami would be also one option.}

\section{Conclusion}
Commencing from the observation that multimodal hate-meme classifiers exhibit poor generalization to other datasets, we demonstrate that comparable cross-domain performance can be achieved by disregarding the image segment and concentrating solely on the text. Furthermore, we reveal that text classifiers exhibit improved performance when incorporating image content into the text classifier through image captioning. Intriguingly, the introduction of captions generated from meme images to the hate meme classifier leads to a deterioration in performance. The insights obtained from the analysis of modality-specific contributions, along with the diminishing effect of including captions, indicate that current multimodal models are primarily focused on finding alignment between image and text tokens at a concrete level. The addition of captions generated by other multimodal models misdirects attention to those low-level alignments, whereas text-image alignment in hate text classifiers typically occurs at a more abstract (metaphorical) level. It is evident that the meaning of the image could be extracted and incorporated at a higher level, where current models and training regimes fall short in addressing this issue. Additionally, our evaluation on a newly established confounder dataset underscores superior performance on text confounders as opposed to image confounders. These findings strongly support the assertion that the image component of multimodal hate meme classifiers exhibits limited transferability, with the generalization capabilities primarily dependent on the text component of the meme.

\section{Limitations}
Employing a proprietary API such as Perspective API introduces challenges to reproducibility. Nonetheless, we mitigate this limitation by training our own BERT classifier, offering a comparably high-performing and fully reproducible alternative. In our approach, we consciously restrict ourselves to a single multimodal classifier, chosen for its high efficiency in general, for both the explanation phase and confounder study. However, consistency across the results enhances the viability of our analysis. There is a lack of propositions about questions such as why \textsc{MAMI} models are different than others not affected at all from caption inclusion, or what makes \textsc{HARMEME} models easier than \textsc{FB}'s (as shown in Table~\ref{rivals}). 
In this study, our focus has been on employing existing models that have been utilized or proposed for the Hateful Meme Classification task. Nevertheless, it is also worthwhile to acknowledge the study like multimodal gate method introduced by \cite{Arevalo2020}. Such a method proposes a systematic control over the contributions of modalities through a multimodal gate mechanism. We also believe that adopting such an approach could offer insights into several aspects, including (i) potential enhancements in hate meme detection, (ii) investigating whether the challenges stem from insufficient attention to the visual modality, and (iii) understanding if, even with increased attention to the visual modality, models might still concentrate on less relevant aspects of inputs, proving counterproductive for meme comprehension. Arguably, considering the recent progress in pre-trained large language models (PLMs) with the ability to analyze multimodal data, exemplified by MiniGPT-4 \cite{zhu2023minigpt4}, they could be contemplated for inclusion in the study. Nevertheless, challenges such as hallucination \cite{li-etal-2023-evaluating}, mainly stemming from their longer average response length, pose concerns that we believe may have implications for tasks like hate meme detection.

\section{Ethics Statement}
Predicting whether a meme is hateful or not might infringe on the fundamental right of free speech if the prediction is used by a government or service provider to remove the post or block the posting user.
If viewed from this perspective, it might be good news that the technology --as we show in this paper-- barely works.
On the other hand, not addressing hate speech would give further rise to possible discrimination, making it a problem for equal participation in any society. In terms of carbon emission, we conducted experiments primarily on GPUs to assess their resilience and develop countermeasure models. Using a private infrastructure with a carbon efficiency of 0.432 kgCO$_2$eq/kWh, we performed 120 hours of computation on 24 GB memory size Quadro RTX 6000 GPU. The total estimated emissions were 15.55 kgCO$_2$eq, with no direct offset \cite{lacoste2019quantifying}.

\section*{Acknowledgments}
This work was conducted at CATALPA -- Center for Advanced Technology-Assisted Learning and Predictive Analytics of the FernUniversität in Hagen, Germany. The fourth author acknowledges financial support by the project ``SAIL: SustAInable Life-cycle of Intelligent Socio-Technical Systems'' (Grant ID NW21-059A), which is funded by the program ``Netzwerke 2021'' of the Ministry of Culture and Science of the State of Northrhine Westphalia, Germany.

\bibliography{custom}
\appendix
\section{List of Controversial Figures/Celebrities}
\label{celebrityList}
Table~\ref{figurelist} presents the comprehensive list of controversial figures and celebrities under consideration for the compilation of the confounder dataset.

\begin{table}[t]
\centering
\small
	\begin{tabular}{l}
		\toprule
		 \bf Controversial Figures/Celebrities \\
		\midrule
            Adolf Hitler \\
Anne Frank \\
Joseph Goebbels \\
Donald Trump \\
Nana Addo Dankwa Akufo-Addo \\
Barack Obama \\
Abu Bakr Al-Baghdadi \\
Joe Biden \\
Osama Bin Laden \\
King Charles \\
Prince Harry \\
Bill Clinton \\
Bill Cosby \\
BillGates \\
Chris Evans \\
James Franco \\
Pauline Hanson \\
Hassan Rouhani \\
Kamala Harris \\
Kevin Hart \\
George W. Bush \\
Hillary Clinton \\
Hulk Hogan \\
Stephen Hawking \\
Martin Luther King Jr. \\
Vince McMahon \\
Colin Koepernick \\
Melania Trump \\
Michelle Obama \\
Nadeschda Andrejewna Tolokonnikowa \\
Wladimir Putin \\
Ilhan Omar \\
Mike Pence \\
Bridget Powers \\
Pope Francis \\
Will Smith \\
Greta Thunberg \\
Justin Trudeau \\
Stevie Wonder \\
Darryl Worley \\
Caitlyn Jenner \\
Conchita Wurst \\
Mark Zuckerberg \\
		\bottomrule
	\end{tabular}
	\caption{Illustrated the list of controversial figures and celebrities used in confounder dataset.}
	\label{figurelist}
\end{table}

\section{Confounder Dataset - Instructions}
\label{annotationProcess}
\begin{figure*}
  \begin{center}
  \resizebox{\textwidth}{!}{%
\includegraphics[]{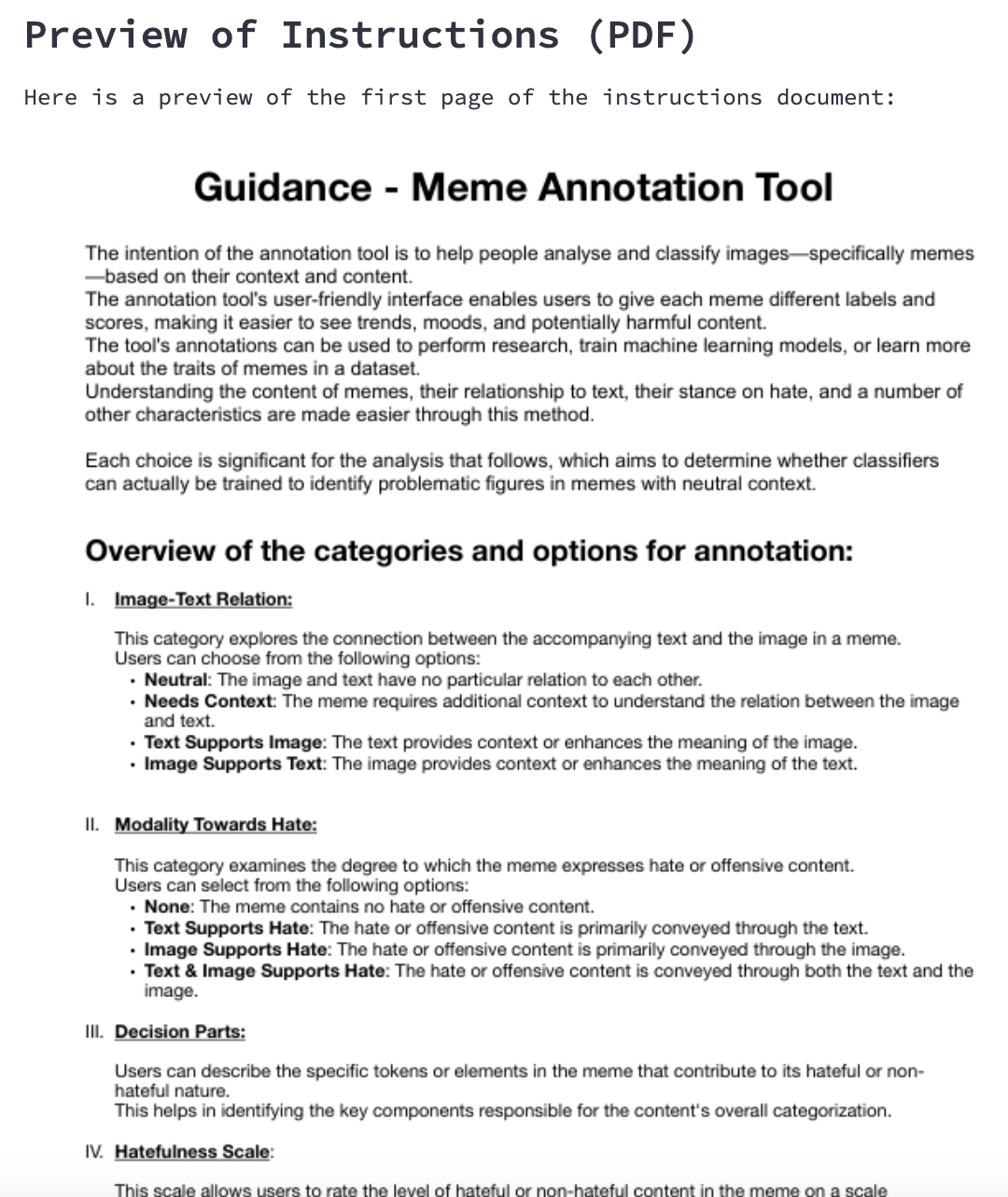}
}
        \end{center}
        \caption{Pdf preview of instruction manual.}

        \label{instructions}
\end{figure*}

Figure~\ref{instructions} illustrate the instructions manual provided to each annotator before starting the annotation process.

\end{document}